\pdfoutput=1

\documentclass[11pt]{article}
\usepackage{makecell}

\usepackage[]{acl}
\usepackage{acronym}
\usepackage{times}
\usepackage{latexsym}
\usepackage{array}
\usepackage{longtable}
\usepackage{tabularx}
\usepackage{adjustbox}
\usepackage{booktabs} 
\usepackage{listings}
\usepackage{amsmath}
\usepackage{amsfonts}
\usepackage[T1]{fontenc}
\usepackage{multirow}
\usepackage[utf8]{inputenc}

\usepackage{microtype}

\usepackage{inconsolata}

\usepackage{graphicx}

\usepackage{xcolor} 

\usepackage{listings}
\usepackage{xcolor}  

\lstdefinestyle{regex}{
    backgroundcolor=\color{white},   
    basicstyle=\ttfamily\small,           
    keywordstyle=\color{blue},             
    commentstyle=\color{gray},            
    stringstyle=\color{red},              
    morekeywords={^, <, >, \{, \}, (, )}, 
    showstringspaces=false,               
    breaklines=true,                      
    breakatwhitespace=false,              
    breakautoindent=true,                 
    captionpos=b,                         
    numbers=none                          
}

\title{A Multi-Stage Workflow for the Review of Marketing Content with Reasoning Large Language Models}

\author{Alberto Purpura\\
  AI Foundations, Capital One
  \texttt{alberto.purpura@capitalone.com} \\\And
  Emily Chen \\
  AI Foundations, Capital One
  \texttt{emily.chen2@capitalone.com} \\\And
  Swapnil Shinde \\
  AI Foundations, Capital One
  \texttt{swapnil.shinde2@capitalone.com} \\}

\author{Alberto Purpura, Emily Chen, Swapnil Shinde\\
  Capital One, AI Foundations \\
  \texttt{\{alberto.purpura, emily.chen2, swapnil.shinde2}
}
  
\begin{document}
\maketitle
\begin{abstract}
Reasoning Large Language Models (LLMs) have shown promising results when tasked with solving complex problems. In this paper, we propose and evaluate a multi-stage workflow that leverages the capabilities of fine-tuned reasoning LLMs to assist in the review process of marketing content, making sure they comply with a given list of requirements. The contributions of this paper are the following: (i) we present a novel approach -- that does not rely on any external knowledge representation -- for the automatic identification of compliance issues in textual content; (ii) compare the effectiveness of different fine-tuning strategies like Supervised Fine-Tuning (SFT) and Group Relative Policy Optimization (GRPO) in training models to solve this problem; (iii) we evaluate the effectiveness of training small LLMs to generate reasoning tokens before providing their final response; (iv) we evaluate how the choice and combinations of different reward functions affects the performance of a model trained with GRPO.
\end{abstract}
\newacro{LLM}{Large Language Model}
\newacro{LLMs}{Large Language Models}
\newacro{GenAI}{Generative AI}
\newacro{GDPR}{General Data Protection Regulation}
\newacro{SFT}{Supervised Fine-Tuning}
\newacro{GRPO}{Group Relative Policy Optimization}
\newacro{MAP}{Mean Average Precision}
\newacro{nDCG}{normalized Discounted Cumulative Gain}
\newacro{TPR}{True Positives Rate}
\newacro{FPR}{False Positives Rate}
\newacro{BLEU}{BiLingual Evaluation Understudy}
\newacro{NER}{Named-Entity Recognition}
\begin{figure*}[h]
    \centering
    \includegraphics[width=\linewidth]{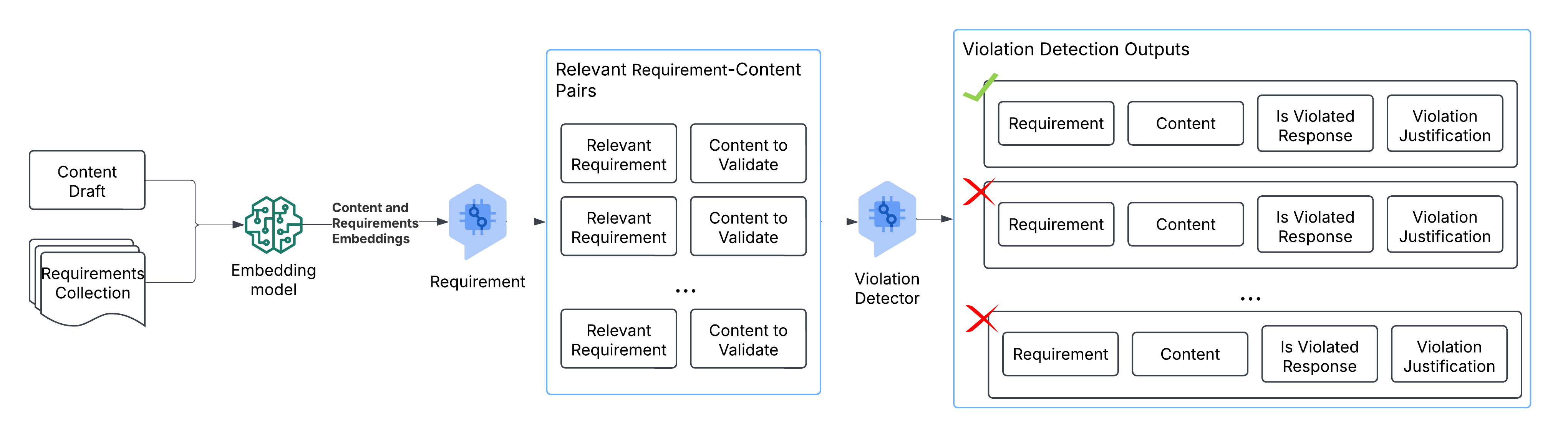}
    \caption{Architecture of the proposed multi-stage workflow for content validation: (i) the content to validate is compared to the list of available requirements and a subset of relevant requirements that may apply to this type of content is selected for the validation stage; (ii) the content is checked against each selected requirement and a boolean value is produced to indicate whether the content passes or fails the validation check, together with a justification for each detected violation.}
    \label{fig:arch_schema}
\end{figure*}

\section{Introduction}

The creation and dissemination of marketing content are pivotal for modern businesses. However, ensuring that this content adheres to complex style and compliance guidelines poses a significant challenge. Traditional manual review processes are often time-consuming, error-prone, and struggle to scale with the increasing volume of digital marketing materials and personalization. This challenge is further exacerbated by the dynamic nature of publication standards and internal compliance policies. 

Recent advancements in the field of \ac{GenAI} have opened new avenues for automating and assisting in various content review tasks, including legal document review and compliance checking. However, reasoning over how intricate requirements apply to specific marketing content remains a complex problem. To address this, we propose a novel multi-stage workflow that leverages the capabilities of a reasoning \ac{LLM} to automate the legal review process of marketing content. This workflow is designed to streamline compliance checks, reduce manual effort, and enhance the accuracy of legal reviews. At the same time, we thoroughly evaluate different training strategies and reward functions to train \ac{LLM}s to tackle the violation detection problem.


The contributions of this paper are the following: (i) we present a novel approach relying on \ac{LLM}s for the automatic identification of compliance issues in textual content that does not rely on any external knowledge representation; (ii) compare the effectiveness of different fine-tuning strategies like \ac{SFT} and \ac{GRPO} \cite{shao2024deepseekmath} in training models able to reason over how a requirement may apply to some marketing content; (iii) we evaluate the effectiveness of \ac{SFT} and \ac{GRPO} for training small \ac{LLM}s generating \textit{reasoning tokens} -- i.e. text explaining a model's reasoning to solve the violation detection problem -- before generating their final response and whether generating reasoning text in this context improves the performance of the model; (iv) we evaluate how the choice of different reward functions affects the performance of a reasoning model trained with \ac{GRPO}. 



\section{Related Work}

The use of \ac{LLM}s in content validation has seen significant growth in the past few years, with applications ranging from \ac{GDPR}\footnote{\url{https://gdpr-info.eu/}} compliance checking \cite{amaral2021ai} to building regulation analysis \cite{chen2024automated}. Existing research works generally approach the problem with a two stage solution. First they extract structured information from the documents to validate and/or reference legal guidelines. Then, they compare the two relying on their respective precomputed representations. 
\citet{amaral2021ai} were among the first researchers to explore AI-enabled automation for completeness checking of privacy policies, focusing \ac{GDPR} regulations. This approach heavily relies on an ontology and \ac{NER} models to parse and compare the content to verify and \ac{GDPR} regulations.
With the growth in popularity of \ac{LLM}s, we observe an increase in the number of approaches relying less on auxiliary information representation structures or human input. For example, \citet{hassani2024enhancing} proposes a multi-step approach to perform compliance checking of documents. Their approach requires a preliminary step of regulatory content parsing and classification with respect to a given set of categories. The content to be analyzed is then chunked and compared to the extracted information through prompting.
In \cite{chen2024automated}, the authors demonstrate another application of \ac{LLM}s combined with ontologies for compliance checks of building codes, highlighting the benefits of structured knowledge representation while still working with \ac{LLM}s. Similarly, \citet{la2024safeguarding} propose \ac{LLM} agents for automating community rule compliance in decentralized social media, showcasing the adaptability of these models to various compliance scenarios.
Here, we explore a further simplification of this class of approaches, greatly reducing the complexity of the compliance checking pipeline by shifting this load on reasoning \ac{LLM}s. \citet{shao2024deepseekmath} recently showed how fine-tuning reasoning \ac{LLM}s with reinforcement learning strategies improved the performance of \ac{LLM}s when answering mathematical questions. This technique has been validated by other researchers in different fields \cite{pan2025medvlm} as showing performance improvements in complex tasks. In this paper, we apply a similar fine-tuning strategy -- with different reward functions -- and compare it to traditional \ac{SFT} for the task of compliance checking of marketing content.

Our work distinguishes itself from the aforementioned approaches in different ways. First, while several of the existing solutions for content validation consider sentences or paragraphs as the base unit for content validation, we consider -- thanks to the recent increases in context size of modern \ac{LLM}s -- an entire document. Second, since we do not rely on any structure or ontology to represent compliance constraints. Third, our system can adjust to a dynamic collection of constantly changing regulations without human intervention. Fourth, we show how to leverage small \ac{LLM}s that rely on reasoning tokens to analyze how a regulation may apply to some content and to identify a violation. Finally, we include a thorough evaluation of different fine-tuning approaches and provide an interpretation of how \ac{GRPO} benefits from the inclusion of the \ac{BLEU} Score as a reward function.


\section{Data and Methods}
\label{sec:methods}
\subsection{Proposed Approach}

\begin{figure*}[h]
    \centering
    \includegraphics[width=0.8\linewidth]{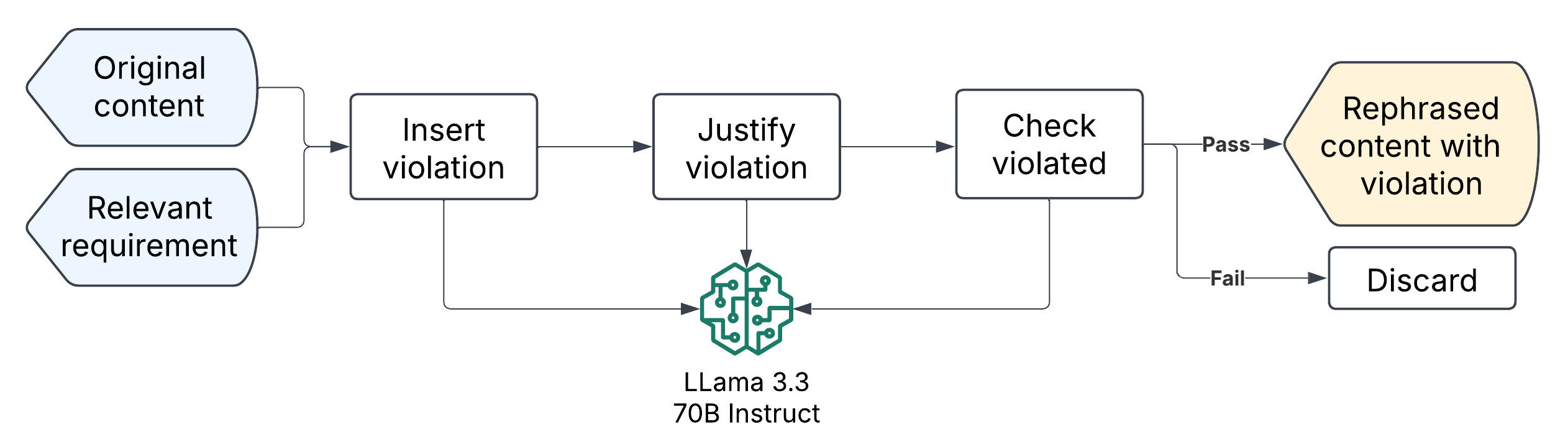}
    \caption{Workflow for the generation of violated content drafts.}
    \label{fig:synthetic_violation_insertion}
\end{figure*}

\begin{figure*}[h]
    \centering
    \includegraphics[width=0.8\linewidth]{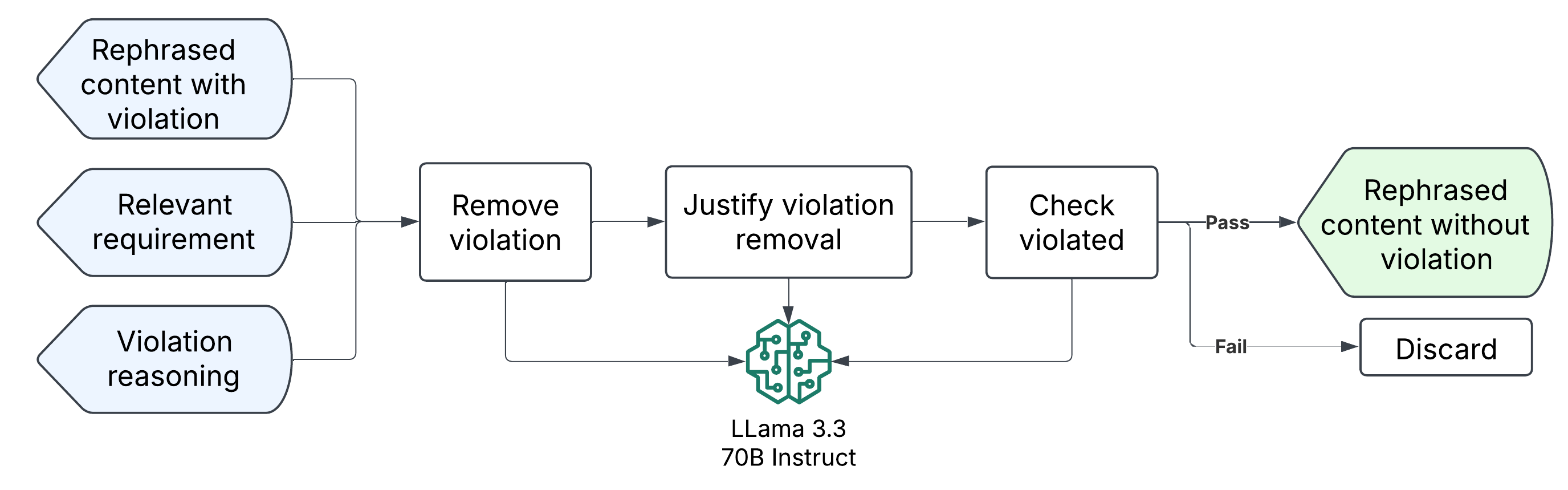}
    \caption{Workflow for the generation of content drafts without violations.}
    \label{fig:synthetic_violation_removal}
\end{figure*}

The proposed workflow is depicted in Figure \ref{fig:arch_schema} and is made up of two stages (i) retrieval and (ii) violation detection. The first stage is responsible for matching relevant requirements to the content draft to evaluate, the second component checks whether the content draft violates (or not) each of the relevant requirements. More details on these components are provided below.

\textbf{Retrieval Stage.} The primary objective of the retrieval stage is to reduce the computational cost and execution time of the entire content validation pipeline. This is achieved by identifying a minimal subset of relevant requirements from a potentially extensive collection, thereby avoiding exhaustive comparisons against the entire regulatory corpus.
We leverage the same text embedding model from \texttt{SentenceTransformers}~\footnote{\url{https://sbert.net/}} to compute dense vector representations for a marketing content draft and requirement texts. 
We fine-tune the embedding model with a cosine similarity loss maximizing the similarity of representations of related requirements and marketing contents. Our relevance labels for each content-requirement pair are binary values generated with \texttt{Llama 3.3 70B-Instruct}. More details on the synthetic data generation process that we follow is available in Section \ref{sec:dataset}.
After fine-tuning this model, we compute the cosine similarity of each pair and use this score to rank them. 

To decide how many results to pass to the next stage of the pipeline, we compute the minimum number of requirements $k$ to return so that, on average, 80\% of our requirements list are associated with a recall of 0.8 or higher. This value also allows us to compute the pipeline efficiency gain achieved thanks to this component. Given a list of $n$ requirements, we define the pipeline efficiency gain measure as $\frac{n-k}{n}$ i.e., the number of requirements over the total that we are not processing in the second stage of the pipeline over the total. We express this value in our evaluation section as a percentage.
Both marketing content and regulatory documents are typically composed of multiple paragraphs. To facilitate efficient retrieval, our team has summarized complex requirements into more concise texts, averaging 176 words each.~\footnote{As an alternative solution -- beyond the scope of this paper -- \ac{LLM}s may be used to identify and paraphrase different requirements from a longer document.}
If the length of either the content draft to analyze or the requirement text is larger than the context length of the text embedding model we are using, we truncate it -- based on our evaluation results reported in Section \ref{sec:results}, this operation does not affect performance results. 

\textbf{Violation Detection Stage.} Following the retrieval stage, the violation detection stage has the role of determining whether some content violates a given requirement and explaining the reasons behind the violation. To implement this functional block we rely on a fine-tuned \ac{LLM} that we train to generate structured output text in the format \texttt{<think>.*</think>} \texttt{<answer>(True|False)</answer>}. The content between the \texttt{<think>.*</think>} tags represents the model's reasoning while the one enclosed between the \texttt{<answer>(True|False)</answer>} tags indicates whether the content violates the provided requirement (True) or not (False). The input to this model is a prompt containing the text of the content draft to verify and the text of a requirement it should comply with. This content is up to 3000 tokens long, while we limit length of the \ac{LLM} response to 500 tokens. For this reason, due to technical limitations, we are not able to perform full parameter fine-tuning of all \ac{LLM} weights. Instead, we take a more efficient approach and fine-tune a LoRA adapter \cite{hu2022lora}.

We experiment with different fine-tuning strategies of a LoRA adapter with \ac{GRPO} \cite{shao2024deepseekmath} and \ac{SFT} -- more technical details on the  LoRA adapter, the \ac{LLM}s we fine-tune and the parameters we use are provided in Section \ref{sec:results}. Recent research works \cite{guo2025deepseek} proved the advantages of reinforcement learning -- \ac{GRPO} in particular -- when training \ac{LLM}s that need to leverage reasoning capabilities to solve math problems. Deciding how a certain requirement applies to some content is not a trivial task even for humans and requires extensive reasoning. For this reason, we choose to experiment with \ac{GRPO} fine-tuning to determine whether this training strategy can improve the reasoning abilities of a model and its performance.
We also conduct an ablation study to assess whether requiring the generation of reasoning text in a response improves the performance of a model. The reward functions that we employ (in different combinations) to train our \ac{LLM} with \ac{GRPO} are the following: (i) response format reward, which is equal to 1 when the response from the model matches the format \texttt{<think>.*</think>} \texttt{<answer>(True|False)</answer>} -- or just \texttt{<answer>(True|False)</answer>} in our experiments without reasoning tokens -- and 0 otherwise; (ii) accuracy reward -- i.e. the accuracy of the model in detecting whether a content violates a requirement or not -- this is equal to 1 each time the boolean prediction in the generated text matches our ground truth and 0 otherwise;   and (iii) reasoning text similarity with the ground truth. To compare the texts, we rely on the \ac{BLEU} Score \cite{papineni2002bleu}. This gives us a measurement of the shared word n-grams in the reasoning text generated by the model and the reasoning text available in our ground truth. Note that even though this function is not directly optimizable in the \ac{SFT} setting, \ac{GRPO} allows us to use an arbitrary function as a reward during fine-tuning.
The proposed reward functions can be formally described by the following expressions:
\begin{equation}
r_{\text{format}} = 
\begin{cases}
1, & \text{if response matches regexp} \\
0, & \text{otherwise}
\end{cases}
\label{eq:format}
\end{equation}

\vspace{-1em}

\begin{equation}
r_{\text{accuracy}} = 
\begin{cases}
1, & \text{if } \text{prediction} = \text{ground\_truth} \\
0, & \text{otherwise}
\end{cases}
\label{eq:acc}
\end{equation}

\vspace{-1em}

\begin{multline}
r_{\text{reasoning}} = \text{BLEU}(\text{generated\_reasoning}, \\
\text{ground\_truth\_reasoning})
\label{eq:bleu}
\end{multline}

For what concerns \ac{SFT}, we train the model using the standard cross-entropy loss to compare the generated text -- with and without the reasoning section depending on the experimental setting -- to the ground truth response available in our training dataset.

\subsection{Synthetic Dataset Creation}
\label{sec:dataset}
To facilitate the development and evaluation of our approach on this task, we have constructed a comprehensive synthetic dataset. This dataset encompasses a wide range of marketing content drafts with applicable marketing requirements, enabling the training and evaluation of reasoning \ac{LLM} systems. More in detail, we synthetically generate: (i) a retrieval dataset and (ii) a violation detection dataset. 
These datasets allow us to simulate a real-world scenario where \ac{LLM}s are employed to assist marketing teams, providing feedback on the requirements directly to creators. We are unable to share this data due to its sensitive nature but we include below an example of a marketing content draft and of a requirement that our workflow was designed to verify.
\noindent

\textbf{Example Marketing Content Draft:}
\noindent
\textit{"Open a new 'Advantage Plus' checking account and enjoy benefits like free online transfers, mobile check deposit, and no monthly maintenance fees for the first six months. Manage your finances with ease and convenience. Visit any branch or apply online today. Terms and conditions apply. Disclosures [...]"}

\noindent
\textbf{Example Requirement:}
\noindent
\textit{"All marketing materials for deposit accounts must clearly and conspicuously disclose any fees associated with the account. If a fee is waived for a promotional period, this must be explicitly stated, along with the date the fee will go into effect. The terms and conditions governing the account, including a complete fee schedule, must be readily accessible to the customer (e.g., via a link on a website, or available upon request at a branch)."}

\noindent


\textbf{Retrieval Dataset.} Given a collection of 680 marketing emails (queries) and 460 requirements (documents), we generate relevance labels for a subset of the content-requirement pairs following the following process: (i) for each content draft, we consider the top 100 most similar requirements -- in terms of cosine similarity -- relying on the vector representations provided by the \texttt{all-mpnet-base-v2} model \cite{reimers-2019-sentence-bert}; (ii) we then create relevance labels for each of these pairs by prompting the \texttt{Llama 3.3 70B-Instruct} model \cite{grattafiori2024llama}. In Table \ref{tab:retriever_dataset} we report some descriptive statistics of our retrieval dataset. Our train and test datasets contain disjoint sets of queries (content drafts) while they share the same collection of documents (requirements).






\begin{table}[h]
\centering
\adjustbox{width=\linewidth}{
\begin{tabular}{lr}
\toprule
\textbf{Statistic} & \textbf{Value} \\ 
\midrule
\textbf{Full Dataset}\\
Number of distinct Content Drafts (queries) & 680 \\
Number of distinct Requirements (documents) & 460 \\ 
\midrule
\textbf{Test Dataset}\\
Number of Content Drafts (queries) &  322 (63.88\%) \\
Avg. Relevant Requirements per Content Draft  &  9.93\\
\midrule
\textbf{Train Dataset}\\
Number of Content Drafts (queries)  &  358 (36.12\%) \\
Avg. Relevant Requirements per Content Draft &  11.36\\

\bottomrule
\end{tabular}
}
\caption{Retrieval stage dataset descriptive statistic.}
\label{tab:retriever_dataset}
\end{table}

\begin{figure}[ht!]
    \centering
    \includegraphics[width=\columnwidth]{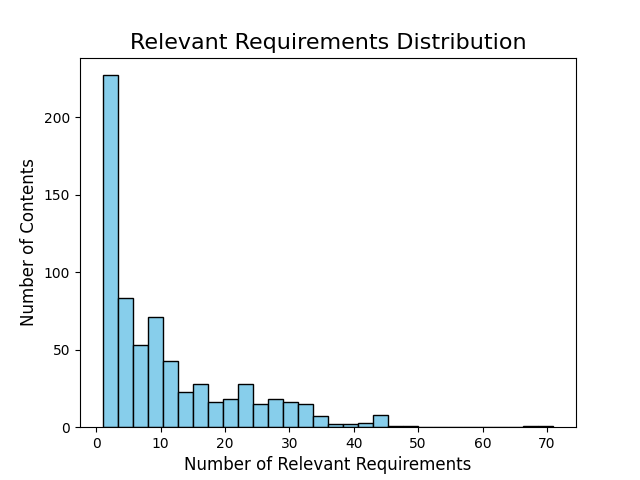}
    \caption{Number of relevant requirements identified for each content draft. 
    }
    
    \label{fig:density_plot}
    \vspace{-1em}
\end{figure}

In Figure \ref{fig:density_plot}, we report the number of relevant requirements we identified for each content draft. From the chart, we observe that the majority of content drafts are matched with 1 to 2 relevant requirements while the number of content drafts associated to more relevant requirements decreases exponentially. 

 
\textbf{Violation Detection Dataset.} 
The violation detection dataset we consider for the training and evaluation of the proposed violation detection component has been generated using the workflows depicted in Figures \ref{fig:synthetic_violation_insertion} and \ref{fig:synthetic_violation_removal}. The objective of both workflows is to have a dataset with synthetic marketing content drafts, a requirement that is relevant and a justification as to why or why not the requirement is respected by the content. 
The first workflow used is shown in Figure \ref{fig:synthetic_violation_insertion}, and it produces marketing content drafts that violate a given requirement. Starting from a piece of content with no violations, the workflow first inserts the violation to a requirement deemed relevant in the previous dataset. The prompt instructs the model to rephrase the content to violate a given requirement, making sure to change as little as possible from the original content i.e., preserving the product description, content structure, style and length.
Then, in a separate call to the same \ac{LLM}, we ask for a justification as to why the newly rephrased content violates the requirement.
Finally, we call the \ac{LLM} to confirm that the rephrased text is violated.

Using the results of the first workflow, we use the one in Figure \ref{fig:synthetic_violation_removal} to remove the violations inserted. We only use the results from the first workflow that are confirmed by the \texttt{Check violated} node. The workflow to remove a violation to a content receives in input the violated content, the requirement it violates and the justification of why it is violated. Its first node prompts an \ac{LLM} to remove the violation from the text, without changing the style, promotional message, structure and length. Similar to the previous workflow, we generate at the next stage the motivations of why the content does not violate the given requirement. Generating this reasoning text at a separate stage is crucial to avoid confirmation biases from the model. 
We also use the same process as in the previous \texttt{Check violation} node to ensure that the violation was successfully removed. After filtering both datasets based on the results of the \texttt{Check violation} node, we combine them to create the validation stage dataset, and report the characteristics in Table \ref{tab:validation_dataset}.

Following this process, we ensure that the violation detection model will be trained on content that definitely violates (or complies with) given requirements. We also ensure that the style of the content is uniform and does not contain artifacts that would allow the model to easily learn shortcuts to achieve its goal. Finally, we balance both the training and test datasets we generated by undersampling the items from the most frequent class.

\begin{table}[h]
\centering
\adjustbox{width=\linewidth}{
\begin{tabular}{lr}
\toprule
\textbf{Statistic} & \textbf{Value} \\ 
\midrule
    \textbf{Full Dataset}\\
        Number of Content Drafts & 12163\\ 
        Number of Requirements & 267 \\
\midrule
    \textbf{Test Dataset}\\
        Number of Content-Requirement Pairs &  4730\\
        Violated &  2162\\ 
        Not Violated &  2162\\
\midrule
    \textbf{Train Dataset}\\
        Number of Content-Requirement Pairs &  7842\\
        Violated &  3044 \\ 
        Not Violated &  3044\\
\bottomrule
\end{tabular}
}
\caption{Violation stage dataset descriptive statistic.} 
\label{tab:validation_dataset}
\end{table}

\section{Evaluation Results}
In this section, we report our performance evaluation of the retrieval model we train to match content drafts with applicable requirements and of the fine-tuned \ac{LLM}s we employ for violation detection. \footnote{All of our experiments were performed on 8 A100 GPUs with 40GB of memory each.}
\label{sec:results}

\textbf{Retrieval Stage.} In Table \ref{tab:retr_perf_eval}, we report the performance of different variants of the proposed retrieval stage. We consider two retrieval models i.e., \texttt{all-mpnet-base-v2} and \texttt{paraphrase-MiniLM-L6-v2} with supported context lengths of 384 and 128 tokens, respectively. We evaluate the performance of both models before and after fine-tuning them for 5 epochs with a batch size of 16 and learning rate 5e-5 using the \texttt{SentenceTransformers} library with pairwise cosine similarity loss.
\begin{table*}[h!]
\centering
\adjustbox{width=\linewidth}{
\label{tab:multi_turn_attacks}
\begin{tabular}{lrrr}
\toprule
\textbf{Approach} & \textbf{Ranked List Size (k)} & \textbf{Recall@k} & \textbf{Efficiency Gain (\%)}\\
\midrule 
\texttt{all-mpnet-base-v2}  & 88 & 0.8726 & 80.87\% \\
Fine-tuned \texttt{all-mpnet-base-v2} & \textbf{50} & 0.8900 & \textbf{89.13\%} \\
\texttt{paraphrase-MiniLM-L6-v2}  & 337 & 0.8680& 26.74\%\\
Fine-tuned \texttt{paraphrase-MiniLM-L6-v2} & 90 & \textbf{0.8974} & 80.43\% \\ 
\bottomrule
\end{tabular}
}
\caption{Retrieval stage evaluation results. Given the ranked list size $k$ and a maximum ranked list size $n$=460, we computed the efficiency gain measure as $(n-k)/n$.}
\label{tab:retr_perf_eval}
\end{table*}
The performance measures that we consider are (i) minimum ranked list size $k$ that allows to achieve a recall higher than 0.8 -- i.e. that allows to retrieve at least 80\% of the relevant requirements -- for 80\% of the content drafts; (ii) Recall@k, where $k$ depends on the previous calculation; and (iii) the pipeline efficiency gain achieved by trimming each ranked list to the amount indicated by the first metric. These measurements allow us to estimate the impact of the retrieval component on the overall efficiency gain of the pipeline thanks to the inclusion of the retrieval model.
As we can observe from the results in Table \ref{tab:retr_perf_eval}, both models benefit in a similar way from additional fine-tuning on the domain-specific data that we are working with. As expected, we observe the best performance results from the larger embedding model model, \texttt{all-mpnet-base-v2}, while the model that proportionally benefited the most from fine-tuning in terms of increased recall and efficiency gain is \texttt{paraphrase-MiniLM-L6-v2}. The fine-tuned \texttt{all-mpnet-base-v2} model allows us to achieve a recall of 0.89 on a ranked list of 50 items. This allows us to confidently prune the set of requirements to validate with an \ac{LLM} to 50, providing an efficiency gain for the content validation workflow of 89\%.

\textbf{Violation Detection Stage.} In Tables \ref{tab:violation_detection_perf_eval} and \ref{tab:violation_detection_perf_eval_reasoning}, we report the performance results of different fine-tuned models for the violation detection component of the proposed workflow. We fine-tune the \texttt{Llama3.2 1B-Instruct} and \texttt{Llama3.2 3B-Instruct} \cite{grattafiori2024llama} models following the different strategies described in Section \ref{sec:methods}. We performed \ac{GRPO} fine-tuning of a LoRA adapter for 2000 steps, with learning rate 1e-6, batch size 8, number of generations equal to 8 and 4 iterations using the \ac{GRPO} TRL library and different combinations of the reward functions indicated in Equations \ref{eq:format}, \ref{eq:acc} and \ref{eq:bleu}.~\footnote{ \url{https://huggingface.co/docs/trl/v0.16.1/en/grpo_trainer}} For \ac{SFT}, we train the model for 5 epochs, with learning rate 5e-5 and batch size 8 using the TRL library. In both cases, we consider a LoRA adapter of rank 16, alpha parameter equal to 32, dropout of 0.05 and no bias term.~\footnote{ We relied on the LoRA implementation available in the PEFT library \url{https://huggingface.co/docs/peft/en/package_reference/lora}}
\begin{table*}[h!]
\centering
\adjustbox{width=\linewidth}{
\begin{tabular}{lllrrrr}
\toprule
\textbf{Base Model} & \makecell{\textbf{Fine-tuning}\\ \textbf{Algorithm}} & \textbf{Reward Functions} & \textbf{Accuracy} & \textbf{Precision} & \textbf{Recall}  & \makecell{\textbf{Correctly}\\ \textbf{formatted}\textbf{ (\%)}} \\
\midrule

\multirow{1}{*}{Llama 3.2 1B Instruct} & SFT & -- & 0.5071 & 0.5092 & 0.3909 & 98.23\% \\
\multirow{1}{*}{Llama 3.2 3B Instruct} & SFT & -- &  \textbf{0.5319} & \textbf{0.5489} & 0.3574 &  \textbf{99.51\%} \\

\multirow{1}{*}{Llama 3.2 1B Instruct} & GRPO & Format, Accuracy &  0.4913 & 0.4942 & \textbf{0.7428} &  87.37\% \\

\multirow{1}{*}{Llama 3.2 3B Instruct} & GRPO & Format, Accuracy & 0.5103 & 0.5148 & 0.3601 & 81.39\% \\
\bottomrule
\end{tabular}
}
\caption{Violation detection stage evaluation results on \ac{LLM}s that do not generate any reasoning tokens. Accuracy, Precision and Recall have been computed considering the performance over the entire test dataset (micro-averaged).}
\label{tab:violation_detection_perf_eval}
\vspace{-0.5em}
\end{table*}
We evaluate the performance of this component -- considering the problem as a binary classification task -- in terms of Accuracy, Micro-Averaged Precision, Micro-Averaged Recall, \ac{BLEU} Score and percentage of correctly formatted responses.\footnote{When we could not parse one of the responses, we considered the output of the model as ``False'', i.e. no violation detected.} For what concerns the \ac{BLEU} Score, we employ this measure to estimate the similarity between the reasoning text generated by the fine-tuned model and the reasoning text in our ground truth computed using \texttt{Llama 3.2 70B-Instruct} with the workflows described in Section \ref{sec:dataset}. The percentage of correctly formatted responses was computed using either of the following regular expressions, depending on whether the approach was supposed to generate any reasoning tokens or not: \texttt{<think>.*</think>} \texttt{<answer>(True|False)</answer>} 
or just its second part after the \texttt{</think>} tag. 
This structured output format allows us to parse the responses from the models and to reliably evaluate their performance.
Another important consideration on our training and test datasets is that the content draft seen by the models during training are different from the ones we used in our test dataset -- while the legal requirements may overlap, as in a real-world scenario. This allows us to more accurately estimate the generalization ability of the models we fine-tune.
From the performance results reported in Tables \ref{tab:violation_detection_perf_eval} and \ref{tab:violation_detection_perf_eval_reasoning}, we observe different performance patterns related to the training strategies, model sizes, the reliance on reasoning tokens and the chosen combination of \ac{GRPO} reward functions.

\textbf{Training strategies.} We observe different performance patterns in the models trained with \ac{SFT} and \ac{GRPO}. 
From the results reported in Table \ref{tab:violation_detection_perf_eval}, we observe how \ac{SFT} is a better fine-tuning strategy for small \ac{LLM}s that directly generate a response to their prompt without any reasoning tokens. 
The situation changes when we require \ac{LLM}s to generate the reasoning for their final response -- see results in Table \ref{tab:violation_detection_perf_eval_reasoning}. In this case, \ac{GRPO} proves to be a more successful strategy compared to \ac{SFT} across all performance metrics. The performance of \ac{GRPO}-trained models varies highly depending on the model sizes, combination of reward functions employed and whether the model generates any reasoning text as outlined below.

\begin{table*}[h!]
\centering
\adjustbox{width=\linewidth}{
\begin{tabular}{lllrrrrr}
\toprule
\textbf{Base Model} & \makecell{\textbf{Fine-tuning}\\ \textbf{Algorithm}} & \textbf{Reward Functions} & \textbf{Accuracy} & \textbf{Precision} & \textbf{Recall} & \makecell{\textbf{BLEU Score}} & \makecell{\textbf{Correctly}\\ \textbf{formatted}\textbf{ (\%)}} \\
\midrule

\multirow{1}{*}{Llama 3.2 1B Instruct} & SFT (R) & -- & 0.4877 & 0.4892 & 0.5601 & 0.0285 & 14.31\% \\ \midrule
\multirow{1}{*}{Llama 3.2 3B Instruct} & SFT (R) & -- & 0.5220 & 0.5341 & 0.3449 & 0.0629 & 37.25\% \\ \midrule
\multirow{3}{*}{Llama 3.2 1B Instruct} & GRPO (R) & \makecell[l]{Format, Accuracy} & 0.4957 & 0.4944 & 0.3748 & 0.0000 & \textbf{99.15\%} \\
 & GRPO (R) & \makecell[l]{Format, BLEU, Accuracy} & 0.4936 & 0.4909 & 0.3453 & 0.0000 & 98.95\% \\
 & GRPO (R) & \makecell[l]{Format, BLEU} & \textbf{0.5016} & \textbf{0.5009} & \textbf{0.9208} & \textbf{0.0008} & 97.73\% \\ \midrule

\multirow{3}{*}{Llama 3.2 3B Instruct} & GRPO (R) & \makecell[l]{Format, Accuracy} & 0.5471 & 0.5612 & 0.4320 & 0.0104 & \textbf{98.37\%} \\ 
 & GRPO (R) & \makecell[l]{Format, BLEU, Accuracy}  & 0.6110 & 0.5881 & \textbf{0.7411} & 0.0400 & 97.96\% \\
 & GRPO (R) & \makecell[l]{Format, BLEU} & \textbf{0.6651} & \textbf{0.6435} & 0.7401 & \textbf{0.0802} & 94.53\% \\
\bottomrule
\end{tabular}
}
\caption{Violation detection stage evaluation results of reasoning \ac{LLM}s. Accuracy, Precision and Recall have been computed considering the performance over the entire test dataset (micro-averaged).}
\label{tab:violation_detection_perf_eval_reasoning}
\vspace{-1em}
\end{table*}

\textbf{Reasoning tokens.} 
The smaller 1B Llama \ac{LLM} -- trained either with \ac{SFT} or \ac{GRPO} -- is in general not able to take advantage of the additional reasoning tokens to increase its accuracy. 
On the other hand, the larger 3B model shows substantial performance improvements. This is true in particular when using \ac{GRPO} as the training strategy. 
It is worth noting however how the \ac{SFT} trained models learn better than the \ac{GRPO} ones to reuse word n-grams patterns from the training data -- this reflects in sometimes higher \ac{BLEU} Scores reported in the results table but not necessarily in higher Accuracy scores. 
Overall, we observe that larger models are better positioned to take advantage of the generated reasoning text compared to smaller ones, especially when trained with \ac{GRPO}.
 
\textbf{Model sizes.} Considering the evaluation results in Table \ref{tab:violation_detection_perf_eval_reasoning},  the \texttt{Llama 3.2 3B - Instruct} model trained with \ac{GRPO} is the one that benefits the most from generating reasoning tokens and that achieves the highest accuracy. This is likely a result of the better reasoning ability of the model, tied directly to its larger number of parameters. If this trend holds, we expect larger models to benefit even further from this fine-tuning strategy. We observe a similar relative performance improvement when evaluating the 3B model variants trained with \ac{SFT} compared to the 1B ones. On the other hand, the \texttt{Llama 3.2 1B - Instruct} model is unable to correctly reason over this problem and to accurately respond in most of the cases.

\textbf{GRPO Reward Functions Combination.}
We repeat \ac{GRPO} fine-tuning of the 1B and 3B Llama models with different combinations of reward functions -- reported in Eq. \ref{eq:format}, \ref{eq:acc}, \ref{eq:bleu} -- to gauge their impact on the performance of the models. These results are reported in Table \ref{tab:violation_detection_perf_eval_reasoning}. 
When relying on the combination of Format and \ac{BLEU} Score reward functions (Eq. \ref{eq:format} and \ref{eq:bleu}), our model achieves the highest Accuracy, Precision and \ac{BLEU} Score metrics, while also maintaining a high Recall and response format accuracy. 
We also observe that explicitly including an Accuracy reward function -- i.e., rewarding directly the model for a correct prediction -- yields lower performance gains compared to relying on the \ac{BLEU} Score. We suspect this is due to the fact that \ac{LLM}s learn through language patterns better than single ground truth signals. In other words, the reasoning text justifying the desired output of the model 
carries more information than the few tokens linked to the computation of the Accuracy reward -- i.e., the \texttt{True/False} tokens at the end of the response. 
Additionally, while the Accuracy score takes binary values for each of the training examples -- either 1 or 0, averaged across each training batch -- the \ac{BLEU} Score has a wider range of values -- in the [0-1] range -- which allows the model to move slowly but with a more continuous signal towards a better policy.

Overall, we observe that \ac{GRPO} is the fine-tuning strategy yielding the best results on this problem. However, its efficacy is closely tied to the initial model abilities and the chosen combination of reward functions. Intuitively, \ac{GRPO}, like other reinforcement learning approaches, relies on the ability of the base model to generate responses to prompts that are then rewarded by the training strategy. For this reason, a model that is unlikely to generate patterns that may be rewarded from the start is unlikely to benefit much from this training strategy. Vice versa, a smaller model may learn more from an explicit training solution like \ac{SFT} which directly updates the weights to increase the likelihood to generate the correct response tokens. 
\section{Conclusion}

In this paper, we presented a novel multi-stage workflow that leverages the capabilities of reasoning \ac{LLM}s to automate the review process of marketing content. Our approach streamlines the compliance checking pipeline by effectively utilizing \ac{LLM}s to reason about the application of requirements to content, thereby reducing the reliance on complex external knowledge representations. We conducted a comparative analysis of different fine-tuning strategies, specifically \ac{SFT} and \ac{GRPO}, to discern their impact on model performance. Furthermore, we evaluated the influence of generating reasoning tokens as part of the model's output, assessing how it contributes to the accuracy of violation detection. Our experiments demonstrate the potential of \ac{GRPO} fine-tuning as a technique to enhance the reasoning capabilities of \ac{LLM}s, particularly in the context of larger models. The findings of our research also underscore the importance of generating reasoning text, as it plays a crucial role in improving the accuracy and reliability of automated violation detection, especially as the size of the violation detection \ac{LLM} increases. We also highlight the importance of choosing the right combination of reward functions when training a model with \ac{GRPO}. Based on our experiments, we observed the largest performance gains when using reward functions spanning a continuous numerical range and linked to a larger number of reasoning tokens rather than binary values linked to few response tokens.
\newpage

\section*{Limitations}
Our work has a few limitations. First, our approach relies on a synthetic dataset. While the generation process we designed allowed us to create a varied dataset that simulates a real-world scenario, the validity of our results depends on how well this dataset represents real data. Second, our experiments are limited to marketing content and a specific set of compliance requirements. While our approach is designed to be general, its effectiveness may vary across different types of content and compliance rules. Third, our current implementation does not address the automated extraction of compliance requirements. In this work, we assume that these requirements are readily available. However, in a real-world scenario, these requirements may need to be extracted from legal documents, which is a complex task in itself.

\section*{Ethics Statement}
The use of \ac{LLM}s to automate compliance checks in marketing content raises a few ethical considerations. One potential concern is the risk of bias in the \ac{LLM}'s predictions. If the training data contains biases, the \ac{LLM} may perpetuate or even amplify these biases, leading to unfair or discriminatory outcomes. For example, the \ac{LLM} may be more likely to flag content as non-compliant if it uses certain demographic keywords.
Another ethical consideration is the potential for the \ac{LLM} to make errors. While our experiments show promising results, the \ac{LLM} is not perfect and may sometimes make incorrect predictions. This could lead to content being wrongly flagged as non-compliant, which could have negative consequences for the content creator. We envision this tool to be used within a process that mitigates for such risks with human supervision and not as a tool to be employed in an unsupervised fashion. Additionally, we designed the framework to include the reasoning for the outputs of the \ac{LLM}s employed. This allows the users of this tool to interpret the underlying reasoning of the model and to reject it if deemed incorrect.
To further mitigate these risks, it is important to carefully curate the training data to minimize biases, to thoroughly evaluate the \ac{LLM}'s performance.

\bibliography{references}


\end{document}